# Localization, balance and affinity: a stronger multifaceted collaborative salient object detector in remote sensing images

Yakun Xie, Suning Liu, Hongyu Chen, Shaohan Cao, Huixin Zhang, Dejun Feng, Qian Wan, Jun Zhu, Qing Zhu,

*Abstract*—Despite significant advancements in salient object detection (SOD) in optical remote sensing images (ORSI), challenges persist due to the intricate edge structures of ORSIs and the complexity of their contextual relationships. Current deep learning approaches encounter difficulties in accurately identifying boundary features and lack efficiency in collaboratively modeling the foreground and background by leveraging contextual features. To address these challenges, we propose a stronger multifaceted collaborative salient object detector in ORSIs, termed LBA-MCNet, which incorporates aspects of localization, balance, and affinity. The network focus on accurately locating targets, balancing detailed features, and modeling image-level global context information. Specifically, we design the edge feature adaptive balancing and adjusting (EFABA) module for precise edge localization, using edge features to guide attention to boundaries and preserve spatial details. Moreover, we design the global distributed affinity learning (GDAL) module to model global context. It captures global context by generating an affinity map from the encoder's final layer, ensuring effective modeling of global patterns. Additionally, deep supervision during deconvolution further enhances feature representation. Finally, we compared with 28 state-of-the-art approaches on three publicly available datasets. The results clearly demonstrate the superiority of our method. The codes of our method are available at https://github.com/little1bold/LBA-MCNet.

## I. INTRODUCTION

Visual saliency detection aims to simulate the human visual system to recognize salient areas [1], [2]. Building this concept, salient object detection (SOD) focuses on identifying and isolating the most visually salient objects within images. SOD finds applications in various fields, such as photo cropping [3], target tracking [4], dynamic detection [5], [6], and semantic segmentation [7], [8]. Originally applied to natural scene images (NSIs), SOD has now expanded to encompass optical remote sensing images (ORSIs). SOD plays a crucial role in scene classification [9-11], object localization [12], and the dynamic monitoring and analysis of remote sensing images [13], [14]. However, salient object detection in optical remote sensing images (ORSI-SOD) poses greater challenges than salient object detection in natural scene images (NSI-SOD), due to the diversity of scenes, complex backgrounds, and the presence of multiple interfering factors [15], [16].

Current ORSI-SOD methods can be broadly classified into two categories: traditional methods and deep learning (DL) methods. Initially, traditional methods primarily depended on specific handcrafted features to detect salient objects, such as feature fusion [17] and color content [18]. With the expansion of data volume, several machine learning approaches emerged to address the challenges, including support vector machine [19], [20] and regularized random walk ranking [21], [22]. However, the complexity of ORSIs hinders traditional methods from automatically extracting features. Their limited non-linear expressiveness constrains the capture of deep semantic image features, resulting in low detection accuracy for salient objects. As shown in Fig. 1, two traditional methods like DSG [23] and RCRR [21] demonstrate constrained proficiency in extracting features for ORSI-SOD, frequently mistaking background elements for salient objects and introducing considerable noise.

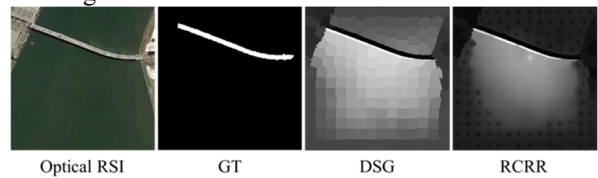

Fig. 1. Illustration of the performance of traditional methods.

The advancements in convolutional neural networks (CNN) have resulted in significant improvements in the accuracy of ORSI-SOD, which faces challenges from complex backgrounds and significant interference [24-27]. Initial research on ORSI-SOD focused on capturing information at various levels within the network structure to reduce interference from cluttered backgrounds [28]. As research progresses, an increasing number of researchers are leveraging the edges of salient objects as auxiliary information [29], [30]. However, unlike NSIs, ORSIs typically feature lower resolution, more complex texture structure and lower quality

This paper was supported by the National Natural Science Foundation of China (Grant Nos. 42301473, 42271424, and 42171397), the Postdoctoral Innovation Talents Support Program (BX20230299), and the China Postdoctoral Science Foundation (2023M742884), the Natural Science Foundation of Sichuan Province (24NSFSC2264), the Key Research and Development Project of Sichuan Province (Grant No. 24ZDYF0633), the University level research project of Chengdu Normal University ((Grant No. CS22XMPY0107). *(Corresponding author: Dejun Feng)*

Y. Xie, H. Chen, S. Liu, S. Cao, D. Feng, J. Zhu are with the Faculty of Geosciences Engineering, Southwest Jiaotong University, Chengdu 610097, China (e-mail: yakunxie@163.com; sakura_ningning@163.com; chy0519@my.swjtu.edu.cn; shcao@my.swjtu.edu.cn; djfeng@swjtu.edu.cn; zhujun@swjtu.edu.cn, zhuqing@swjtu.edu.cn).

H. Zhang is with College of Physics and Engineering, Chengdu Normal University, Chengdu 610041, China (e-mail: 523054@cdnu.edu.cn).

Q. Wan is with the Faculty of Foreign Languages, Southwest Jiaotong University, Chengdu 610097, China (e-mail: wanqian@my.swjtu.edu.cn).

edge display. Most existing methods for acquiring edge information rely on the shallow features output by encoders [31], [32]. During transmission, these shallow features are vulnerable to background noise, resulting in blurred boundary outputs. As shown in Fig. 2, the boundaries of the aircraft detected by these boundary-based methods float within the yellow buffer zone, making it difficult to pinpoint the aircraft boundary. However, ORSIs contain a vast array of target types, and the backgrounds are significantly more complex. The existing methods possess limited capability in responding to the differences and relationships between the foreground and background [33-35]. As illustrated in Fig. 3, the global-local embedding module fails to effectively capture and integrate global and local information, leading to misidentifications such as mistaking roads adjacent to buildings as salient objects. It also exhibits limited discriminative ability between salient objects and backgrounds. Overall, the primary challenges revolve around how to balance detailed features with local semantics across various levels, ensuring each feature layer contains rich semantic information and precise location details, and how to model the image-level global context to encourage an affinity for common characteristics.

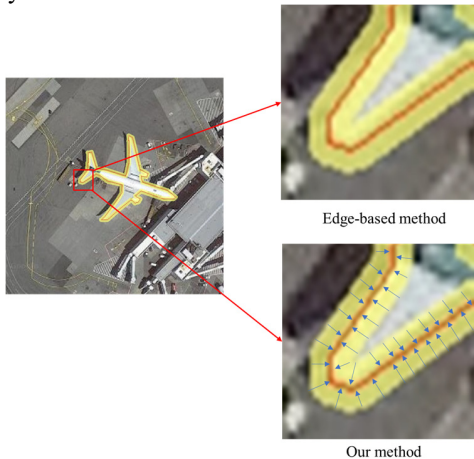

**Fig. 2**. Comparison with edge-based method. The yellow buffer represents the boundary detected by the edge-based method fluctuating within this region. The blue arrow represents that our method pinpoints and gets closer to the edge information.

Given the limitations and complexity of backgrounds in ORSIs, we propose a multifaceted collaborative network that encompasses localization, balance, and affinity components (LBA-MCNet). It accurately identifies targets, balances detailed features, and models global context. Using a transformer structure, the network extracts four layers of global features. For shallow features, the EFABA module captures boundaries and balances details and semantics, improving localization accuracy. For deeper features, the GDAL module models global context, enhancing the network's ability to distinguish targets from the background and improve detection accuracy.

Our primary contributions are outlined as follows:
1) We propose a novel LBA-MCNet. Compared to existing approaches, LBA-MCNet strategically balances boundaries near objects, addressing edge location inaccuracies and enhances the collaborative modeling of background and foreground contexts through affinity learning. Compared with 28 previous state-of-the-art methods on three datasets, LBA-MCNet is significantly advanced.

2) We propose an edge feature adaptive balancing and adjusting (EFABA) module to balance local semantics and spatial details. EFABA augment the localization of salient areas and the representation of spatial details to improve its capability to discern edge cues in fine-grained features.

3) We propose a global distributed affinity learning (GDAL) module to model global semantic clues through both explicit and implicit global affinity learning to enhance the network's capacity to discern salient objects effectively.

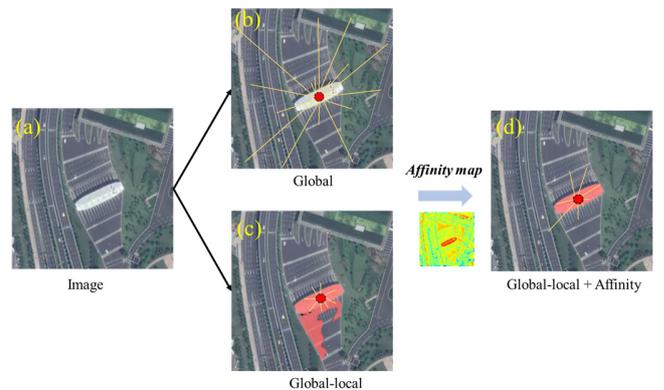

**Fig. 3**. (a) Image. (b) Global. Capturing global information to perceive the overall structure and content of the image. (c) Global-local. Further sensing the information around the building and extracting local features. The building and the surrounding roads are recognized as salient targets. (d) Global-local+Affinity. Besides, the short yellow lines are for schematic display, with the short lines representing local features and the long yellow lines representing global features.

## II. RELATED WORK

In this section, we introduce SOD methods that focus on boundary and context information, respectively.

### A. Boundary-Based SOD

Due to the multiple targets and complex edge textures in ORSIs, salient images frequently display low-quality boundary details. This occurs if fine edges or boundary details of objects are not adequately considered during salient detection. In the quest for high-precision ORSI-SOD, the importance of edge details has grown increasingly significant. Some methods utilize shallow features, which are rich in boundary information, to directly enhance the accuracy of salient object detection by leveraging edges derived from these features [36]. For example, generating boundary information through edges involves monitoring them with boundary labels and emphasizing salient areas [29]. Furthermore, refining the initial saliency map's coarse boundaries using rich edge features leads to the achievement of a more detailed saliency map [31]. However, the coarse boundaries obtained directly from shallow features are often challenged by noise and complex backgrounds. With the advent of various attention mechanisms, some methods have started to process edge information by establishing these mechanisms [37-39]. For instance, spatial attention and channel attention contribute to boundary exploration [39], and the resulting boundary-aware saliency map acts as a cue for implementing edge perception attention mechanisms [32]. However, applying attention mechanisms requires calculating attention weights for the entire feature map, leading to significant computational costs during training. Some research explores lightweight ORSI-SOD possibilities based on semantic matching and edge alignment [40]. Additionally, shallow features typically capture only local information and lack an understanding of the global context and structure of the image. In contrast, deep features often contain rich semantic information. Many methods achieve SOD by fusing different layers' features. For example, combining high-level semantics, low-level details, and edge information for salient object detection [34]. Under the guidance of attention and reverse attention mechanisms, merging high and low-level semantics to progressively refine predictions from coarse to fine [41].

In summary, the aforementioned methods utilize boundary information obtained from shallow features for SOD. Shallow features are limited to local areas and lack deep semantic information, making them less effective in handling noise and complex backgrounds. Introducing attention mechanisms and deep semantic information can significantly improve SOD accuracy, but it also requires considering computational efficiency and the complexity of integrating different information sources. Inspired by these methods, our approach incorporates both spatial and channel attention, effectively integrating them with deep-level information. This allows us to balance detailed features and local semantic information across different levels while accurately locating salient targets.

*B. Context-Based SOD*

ORSIs frequently feature cluttered backgrounds, presenting considerable interference within the background area. Therefore, it's crucial to thoroughly exploit the abundant contextual information present in these images to improve the precision of salient object detection. Presently, numerous methods utilize contextual information alongside encoders to extract features at multiple levels [42]. For example, delving into the contextual information of adjacent features to bolster the context capture capacity of conventional decoder blocks [43], [44]. Enhancing networks' ability in SOD by setting context information as an attention mechanism [45]. The contextual information captured spans both global and local dimensions. Within the network's architecture, the merging of surface-level details with deep semantic insights commonly occurs via global connections, facilitating the encoding and decoding processes at matching scales. A variety of methods have been devised to seamlessly integrate local and global information, significantly improving the ability to capture contextual details [46-49]. This enhanced capability aids in the precise identification of salient objects and background areas. For instance, identifying potential salient objects' locations within features from a global semantic viewpoint [50]. Generating an attention map that integrates local spatial context, global spatial context, and channel context, then applying it to refine encoder features [51].

Most of the methods mentioned previously depend on CNN encoders. The CNN encoders can extract the local features of the image well. However, there are limitations in global information capture, which cannot make full use of global context information. In contrast, transformer encoders stand out for their ability to understand the content of entire scenes, demonstrating strong capabilities in capturing global information. With the introduction of the transformer model, some methods have started to investigate both local and global information leveraging this architecture, tapping into its advanced processing capabilities [52]-[54]. One method employs a self-adaptive spatially labeled Transformer encoder to extract both global and local features. It then integrates these features with global-local semantics and context dependency through specialized decoders, enhancing feature representation and interpretation [55].

To sum up, the above methods explore effective ways to gather global and local contextual information. They achieve this by integrating both global-local and local-global forms of information fusion to enhance the network's discriminative ability for SOD. However, they have not fully explored image-level global context, which poses challenges in handling complex scenarios, especially in ORSIs. Building upon these approaches, our method leverages a Transformer encoder to effectively capture global context information. It adopts a global-local-global pattern, employing explicit and implicit global affinity learning to model image-level global context. This approach strengthens discriminative features, enhancing the network's ability to differentiate between salient objects and backgrounds with varying responses.

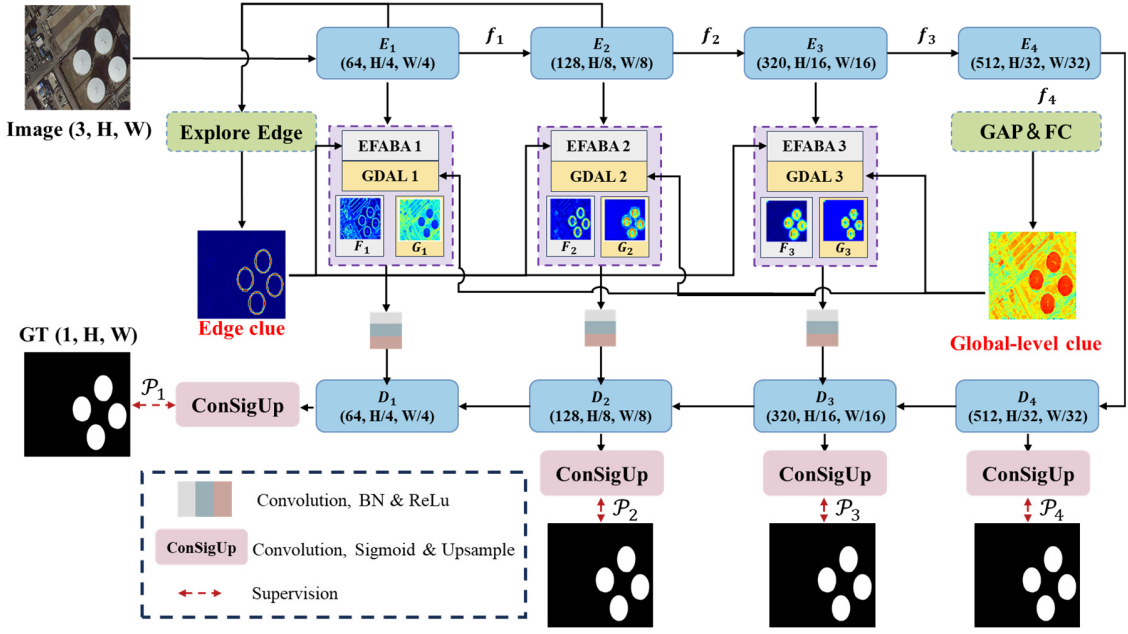

**Fig. 4.** The overall architecture of the proposed LBA-MCNet. $E_1$-$E_4$ represent the process of encoder, $D_1$-$D_4$ represent the process of decoder, $F_1$-$F_3$ represent EFABA results, $G_1$-$G_3$ represent GDAL results.

## III. METHODOLOGY

### A. Network Overview

Fig. 4 depicts the comprehensive architecture of LBA-MCNet, characterized by a symmetrical U-shaped structure. This architecture integrates three core components: the encoder, the EFABA module, and the GDAL module. The Transformer-based encoder facilitates the extraction of features over extended spatial ranges. The EFABA module ensures a balance of local semantics with spatial details, whereas the GDAL module is dedicated to modeling the global context and enhancing feature affinity across the network. Deconvolution processes in the final step refine the feature maps to distinctly highlight salient objects, underscoring the network's refined edge detection. The methodological synergy of location, boundary, and affinity elements forms a complex, collaborative network, markedly enhancing the precision of edge recognition and the network's overall discernment capabilities.

For a single image $i \in \mathbb{R}^{3 \times H \times W}$, we initially utilize the Pyramid Vision Transformer (PVT-V2-B2) to extract multi-layer features, denoted as $f_i \in \mathbb{R}^{C_i \times H_i \times W_i}$. Here, $C_i$ represents the number of channels mapped by the features, for $i=\{1, 2, 3, 4\}$, with the corresponding number of channels for each layer being $\{64, 128, 320, 512\}$. $H_i = (H/2^{i+1})$ denotes the height and $W_i = (W/2^{i+1})$ represents the width. Subsequently, the EFABA and GDAL are applied to the first three layers of features, $\{f_i\}_{i=1}^{3}$, of the encoder. The output feature mappings, denoted as $\{FG_i\}_{i=1}^{3}$, from EFABA and GDAL, are then effectively fused into the decoder, facilitating the synthesis of these advanced features for precise object detection. Ultimately, a sequence of addition fusion and deconvolution operations is performed to reconstruct the resolution of the feature mapping, thereby refining the details and clarity of the final image output. Furthermore, each layer of the decoder outputs a side prediction detection map of varying resolution, $\{\mathcal{P}_i\}_{i=1}^{4}$, using 1x1 convolutions, and undergoes deep supervision. This approach enables the network to jointly detect representations of multi-source information, with $\mathcal{P}_1$ serving as the final output. This strategy ensures that the network leverages the comprehensive information available across various scales and depths for accurate detection.

### B. Edge Feature Adaptive Balance Adjustment Module

Fig. 5 showcases the intricate architecture of EFABA, which includes the edge clue detector (ECD) and the feature adaptive balance adjuster (FABA). Edge clues, characterized as shallow features, are predominantly located in the initial layers of the encoder. The edge information of salient objects significantly improves their localization, thereby enhancing the overall performance of detection mechanisms. Therefore, the integration of edge information into the network's architecture is crucial for optimizing its performance in detecting and localizing objects. Consequently, we implement ECD to encourage the deep network to identify edge clues within fine-grained features. ECD also lays the groundwork for feature balancing across various levels in the subsequent stages, ensuring a harmonized and effective feature representation throughout the network. Leveraging this foundation, the network perceives a richer local context without losing fundamental semantics at any level. Ultimately, fusion features are derived through the feature fusion process, which can be outlined in two main steps:

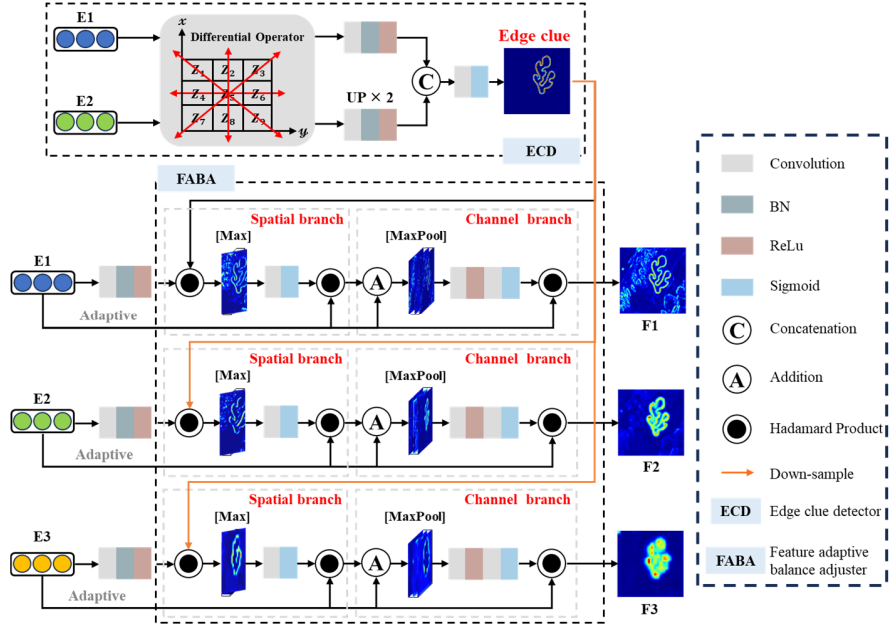

**Fig. 5.** Architecture of the proposed EFABA module. $E_1$-$E_3$ represent the first three layers' features. $F_1$-$F_3$ represent the output of three layers after EFABA module.

*1) Edge clue detector:* Firstly, to capture basic edge features, a first-order differential operator, namely the Sobel operator, is applied to the feature mapping of the first two layers of the encoder, denoted as $\{f_i\}_{i=1}^2$. Specifically, we construct two 3×3 matrices, $\begin{bmatrix} 1 & 0 & -1 \\ 2 & 0 & -2 \\ 1 & 0 & -1 \end{bmatrix}$ and $\begin{bmatrix} 1 & 2 & 1 \\ 0 & 0 & 0 \\ -1 & -2 & -1 \end{bmatrix}$, and use these to perform convolution operations on the original feature mappings of $\{f_i\}_{i=1}^2$, to approximate the grayscale deviations in the horizontal and vertical directions. Subsequently, the gradient magnitude for each point in the feature mapping is calculated, denoted as $\mathcal{g}$. Finally, we acquire the basic edge features, denoted as $\{e_i\}_{i=1}^2$. The entire process can be described as

$$\mathcal{g} = \sqrt{x^2 + y^2} \qquad (1)$$

$$\{e_i\}_{i=1}^2 = \sigma(\mathcal{g}) \otimes \{f_i\}_{i=1}^2 \qquad (2)$$

where x, y represents the grayscale deviations in the horizontal and vertical directions respectively, $\sigma(\cdot)$ represents the Sigmoid activation function, $\otimes$ represents element-by-element multiplication.

Secondly, $\{e_i\}_{i=1}^2$ are input into a 3×3 convolution layer to refine the edge information for the first time. Subsequently, the processed edge information is fused back with $\{f_i\}_{i=1}^2$ through an addition operation. Lastly, the number of feature channels is reduced to 1 using 1×1 convolution block to obtain preliminary fused edge clues $\{e_i^f\}_{i=1}^2$. The process can be expressed as

$$\{e_i^f\}_{i=1}^2 = \mathcal{CBR}_{1\times 1}[\mathcal{C}_{3\times 3}(\{e_i\}_{i=1}^2) \oplus \{f_i\}_{i=1}^2] \qquad (3)$$

where $\mathcal{CBR}_{1\times 1}(\cdot)$ denotes a convolutional layer with a 1×1 kernel, followed by Batch Normalization (BN) and ReLU activation function, $\mathcal{C}_{3\times 3}(\cdot)$ denotes a convolutional layer with a 3×3 kernel, $\oplus$ denotes element-by-element addition.

Finally, to acquire more comprehensive edge clues, the edge features from two layers are further fused. Specifically, the procedure begins by upsampling $e_2^f$ to match the size of $e_1^f$. Next, the edge features from these two layers are input into a 1x1 convolution block, where fusion occurs at the channel level. Ultimately, rich edge attention is achieved by applying the Sigmoid function. The process can be represented as

$$e^{att} = \sigma(\mathcal{C}_{1\times 1}\{\varphi[\mathcal{CBR}_{1\times 1}(e_1^f), \mathcal{CBR}_{1\times 1}(\mathcal{U}(e_2^f))]\}) \qquad (4)$$

where $e^{att}$ denotes the output edge attention, $\varphi(\cdot)$ denotes the concatenate operation, $\mathcal{U}(\cdot)$ denotes the upsampling operation, $\mathcal{C}_{1\times 1}(\cdot)$ denotes a convolutional layer with a 1×1 kernel.

Additionally, to improve the network's representation of multi-source features, the interpolated $e^{att}$ is considered a side output of the network. Then we supervise it deeply through the real boundary.

*2) Feature adaptive balance adjuster:* To adjust the balance between detailed features and local semantic context across various levels, thereby directing the network to emphasize boundary regions for the retention of more complex spatial and geometric details, we firstly use $e^{att}$ as the edge attention to enhance the network's sensitivity to object edges. Meanwhile, we conduct element-by-element multiplication between $e^{att}$ and the feature maps from the first three layers of the encoder, denoted as $\{f_i\}_{i=1}^3$. This step is aimed to minimize the loss of boundary clues during forward propagation. Through the adaptive process of element-wise multiplication between features at different levels, this method is used to fill both high-frequency and low-frequency areas. Subsequently, to further accentuate salient objects while preserving spatial details, attention is directed towards the most prominent feature representations in the channel dimension through the application of a Max operation. After this, we employ a convolution with a larger kernel size of 7 to capture object features from a broader spatial perspective. Following this, the sigmoid function is used to obtain spatial attention, denoted as $\{SAT_i\}_{i=1}^3$. This process is mathematically expressed as follows

$$\{SAT_i\}_{i=1}^3 = \sigma\{C_{7\times7}[\mathcal{M}(e^{att} \otimes \{f_i\}_{i=1}^3)]\} \quad (5)$$

where $\mathcal{M}(\cdot)$ represents the Max operation, $C_{7\times7}(\cdot)$ represents a convolutional layer with a 7×7 kernel.

Subsequently, we multiply and add $\{SAT_i\}_{i=1}^3$ with $\{f_i\}_{i=1}^3$ to achieve spatial alignment. Finally, to minimize noise and accentuate geometric details, we feed the aligned features into channel attention for channel response calibration and multiply with $\{f_i\}_{i=1}^3$ to obtain the final balanced features, denoted as $\{F_i\}_{i=1}^3$. This process can be expressed as

$$\{F_i\}_{i=1}^3 = \{f_i\}_{i=1}^3 \otimes CT[(\{SAT_i\}_{i=1}^3 \oplus 1) \otimes \{f_i\}_{i=1}^3] \quad (6)$$

where $CT(\cdot)$ represents channel attention.

### C. Global Distribution Affinity Learning

Fig.6 illustrates the detailed architecture of GDAL. Global context enhances detection accuracy by ensuring semantic consistency across the image and establishing associations between target objects. Consequently, GDAL is meticulously crafted to model the image-level global context, thereby sharpening the network's ability to distinguish salient objects. This enhancement not only elevates distinctive feature representation but also boosts the network's capacity to differentially respond to foreground and background elements. To achieve this, GDAL models global semantic cues using a dual approach of explicit and implicit global affinity learning, operationalized through two distinct branches:

*1) Explicit affinity learning:* The last layer of encoder rich in semantic knowledge, is crucial for enabling the network to capture image-level global context when utilized effectively. First, the last layer of the encoder, denoted as $\{f_4\}$, undergoes global average pooling followed by a passage through a fully connected layer. This process is designed to extract image-level information, resulting in a feature representation identified as $\{F_4\}_{level}^{img}$. Subsequently, the spatial attention mechanism detailed in Section III-B2 is applied to $\{F_4\}_{level}^{img}$, allowing the model to focus on particular objects or regions. This attention-enriched feature is then propagated to other encoder layers through direct and effective element-wise multiplication. This step facilitates the self-adaptation of global affinity features and leads to the formation of an image-level global context association graph, denoted as $\mathcal{R}$. Finally, the feature association diagram is redistributed across each layer, instructing the model to accentuate global features. The utilization of the Squeeze and Excitation (SE) attention mechanism further sharpens the focus on the most distinctive feature channels. This enhancement bolsters the model's capacity to differentiate between object categories and identify discriminative features more effectively. This process is mathematically expressed as follows

$$\{F_4\}_{level}^{img} = \mathcal{F}[\mathring{A}(\{f_4\})] \quad (7)$$

$$\mathcal{R} = \sigma\{\{f_i\}_{i=1}^3 \otimes \ell[ST(\{F_4\}_{level}^{img})] \quad (8)$$

$$\{f_i^{Exp.}\}_{i=1}^3 = SE(\mathcal{R} \otimes \{f_i\}_{i=1}^3) \quad (9)$$

where $\mathring{A}(\cdot)$ represents global pooling, $\mathcal{F}(\cdot)$ represents fully connected layer, $\ell(\cdot)$ represents sampling operation, $ST(\cdot)$ represents spatial attention, $SE(\cdot)$ represents SE attention, $\{f_i^{Exp.}\}_{i=1}^3$ is the final output of the explicit affinity learning.

*2) Implicit affinity learning:* To capture hierarchical features from images and effectively discern global patterns, our study adopts implicit affinity learning for encoding global information. Similarly utilizing the image-level features $\{F_4\}_{level}^{img}$, we first align its channel dimensions with $\{f_i\}_{i=1}^3$. For simplicity, we assume that after dimension adjustment, $\{f_4\}_{level}^{img} \in \mathbb{R}^{N,C,H,W}$. Then, the feature map is transformed into $\{f_4\}_{level}^{img} \in \mathbb{R}^{N,C,HW}$, whereupon applying the softmax function across the spatial dimension $HW$ generates an attention map, denoted as $\mathcal{M} \in \mathbb{R}^{N,C,HW}$. This process can be described as follows

$$\mathcal{M} = \varpi[\mathcal{T}(\{f_4\}_{level}^{img})] \quad (10)$$

where $\varpi(\cdot)$ represents softmax function and $\mathcal{T}(\cdot)$ represents transpose operation.

Next, $\mathcal{M}$ is transposed, and through matrix multiplication with $\{f_4\}_{level}^{img} \in \mathbb{R}^{N,C,HW}$, we derive the semantic assignment features, denoted as $\{f_4\}_{dis}^{img} \in \mathbb{R}^{N,C,C}$. It's important to note that $\{f_4\}_{dis}^{img}$, being the result of operations on $\{f_4\}$, embodies diverse global representations across its dimensions. Therefore, this method implicitly incorporates a semantic region from the feature map into each layer.

$$\{f_4\}_{dis}^{img} = \mathcal{T}(\mathcal{M}) \odot \{f_4\}_{level}^{img} \quad (11)$$

where $\odot(\cdot)$ represents matrix multiplication.

Furthermore, $\{f_4\}_{dis}^{img}$ is designed to adaptively select affinities with each encoding layer and is reshaped to match the dimensions of each encoding layer. To preserve global semantic integrity, it is then fused with the layer possessing the strongest semantic information, yielding the final implicit affinity feature, denoted as $\{f_i^{Imp.}\}_{i=1}^3 \in \mathbb{R}^{N,C,kH,kW}$. Ultimately, the GDAL's final output, denoted as $\{\mathcal{F}_i^{Exp-Imp}\}_{i=1}^3$, is obtained by integrating features from both explicit and implicit learning pathways.

$${f_i^{Imp.}}_{i=1}^3 = \mathcal{R}(\{f_4\}_{dis}^{img} \odot \{f_i\}_{i=1}^3) \oplus CBR_{1\times1}[\ell(\{f_4\})] \quad (12)$$

$$\{\mathcal{F}_i^{Exp-Imp}\}_{i=1}^3 = CBR_{1\times1}\{CBR_{3\times3}[\varphi(\{f_i^{Imp.}\}_{i=1}^3, \{f_i^{Exp.}\}_{i=1}^3)]\} \quad (13)$$

where $CBR_{3\times3}(\cdot)$ denotes a convolutional layer with a 3×3 kernel.

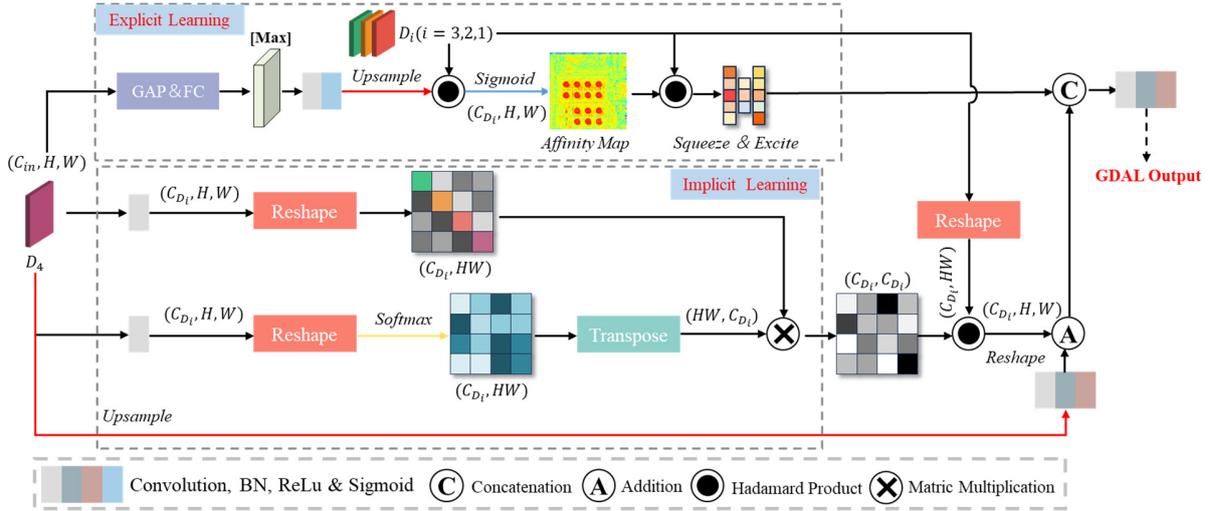

**Fig. 6.** Architecture of the proposed GDAL module. $D_1$-$D_4$ represent the four layers' features.

Explicit affinity learning (EAL) enables the model to focus more purposefully on locations associated with specific objects or regions. On the other hand, Implicit affinity learning (IAL) process more complex, deeper features rich in global information and encode global information representations, thus effectively capturing global patterns in images. Combining the two to model image-level global context improves the network's ability to discriminate salient targets and improves the ability to respond to salient targets that differ from the background.

## IV. EXPERIMENTS

### A. Datasets and Implementation Details

*1) Datasets:* Our method is trained, tested, and evaluated across three publicly accessible ORSI-SOD datasets.

**ORSSD** [28]: The ORSSD dataset is the inaugural publicly available dataset for ORSI-SOD. It consists of 800 optical remote sensing images sourced from Google Earth and various other RSI datasets, each accompanied by manually annotated pixel-level GT images. ORSSD is divided into 600 images for training and 200 images for testing, with each image matched with its corresponding GT image.

**EORSSD** [56]: The EORSSD dataset extends the variety of salient object types and challenging scenarios provided by the ORSSD dataset. It contains 2000 images, each accompanied by their respective pixel-level GT images, with 1400 allocated for training and 600 for testing.

**ORSI-4199** [57]: The ORSI-4199 dataset includes a collection of remote sensing images that span a variety of sizes, scales, and complex backgrounds, offering a more rigorous challenge for SOD. It comprises 4199 images across diverse scenarios, each accompanied by its corresponding pixel-level ground truth (GT) images. The dataset is divided into 2000 images for training and 2199 for testing.

*2) Implementation Details:* All experiments were conducted on a server equipped with an NVIDIA GeForce RTX 3090 24GB GPU. We resized the images to 352×352 pixels and implemented various data augmentation strategies, such as random rotation, cropping, and contrast adjustments, to mitigate overfitting. The Adam optimizer was employed, with a learning rate of 1e-4 and a batch size of 8. The training duration was set to 50 epochs for the ORSSD dataset, 45 epochs for EORSSD, and 35 epochs for ORSI-4199, adjusting the learning rate to 0.1 after every 30 epochs. Fig. 7 illustrates the loss during training of the ORSI4199 dataset. To maintain fairness in our experiments, we obtained the results for other advanced methods using either the provided source codes or saliency maps by their authors. Specifically, for the NSI-SOD method, we retrained using the ORSI-SOD dataset according to the recommended configurations.

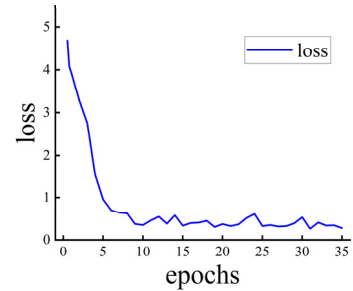

**Fig. 7.** Training loss map for the ORSI4199 dataset.

### B. Comparison with State-of-the-Art Methods

We compare and analyze the method proposed in this study with 28 advanced methods, including two traditional methods, seven NSI-SOD methods, and 19 ORSI-SOD methods.

*1) Quantitative Analysis:* To ensure a comprehensive evaluation of all methodologies, we utilized an array of curves and metrics to facilitate a detailed quantitative analysis.

*a) Curves analysis:* We evaluated the performance of the saliency model using F-measure curves and Precision-Recall (P-R) curves, presented across two rows in Fig. 8. For the F-measure curves, a broader area under the curve indicates superior model performance, while in the P-R curves, proximity to the upper right corner (coordinate (1,1)) denotes better performance. Fig. 8 clearly shows our method outperforming competing approaches across both metrics.

*b) Metrics analysis:* To assess the performance across the three datasets, we utilized three widely recognized SOD metrics: mean absolute error value ($\mathcal{M}$) [58], S-measure($S_\alpha$) [59], maximum F-measure ($F_\beta^{max}$) [60], mean F-measure ($F_\beta^{mean}$), adaptive F-measure ($F_\beta^{adp}$), maximum E-measure ($E_\xi^{max}$) [61], mean E-measure ($E_\xi^{max}$), adaptive E-measure ($E_\xi^{max}$). A lower $\mathcal{M}$ score is indicative of superior performance, whereas higher values of others reflect better performance. The formulas for each evaluation indicator are described below. The comparative results are detailed in Table I, Table II and Table III.

$$\mathcal{M} = \frac{1}{H \times W} \sum_{x=1}^{H} \sum_{y=1}^{W} \|S(x,y) - G(x,y)\| \quad (14)$$

where $\mathcal{M}$ represents the mean absolute error value between the predicted saliency map and GT image. $S$ represents the predicted saliency map, $G$ represents GT image, $H$ and $W$ represent the height and the width of $S$ and $G$ respectively.

$$S_\alpha = \alpha \times S_o + (1-\alpha) \times S_r \quad (15)$$

where $S_\alpha$ represents region-aware and object-aware structural similarity evaluation value, $S_o$ represents the object-aware structural similarity measure, $S_r$ represents the region-aware structural similarity measure, $\alpha$ is set to 0.5.

$$F_\beta = \frac{(1+\beta^2) Precision \times Recall}{\beta^2 Precision + Recall} \quad (16)$$

where $F_\beta$ represents the weighted harmonic of precision and recall, $\beta^2$ is set to 0.3.

$$E_\xi = \frac{1}{W \times H} \sum_{x=1}^{W} \sum_{y=1}^{H} \emptyset_{FM}(x,y) \quad (17)$$

where $E_\xi$ represents binary foreground map evaluation enhanced-alignment measure, $\emptyset_{FM}$ represents the enhanced alignment matrix, H and W are the height and the width of the map, respectively.

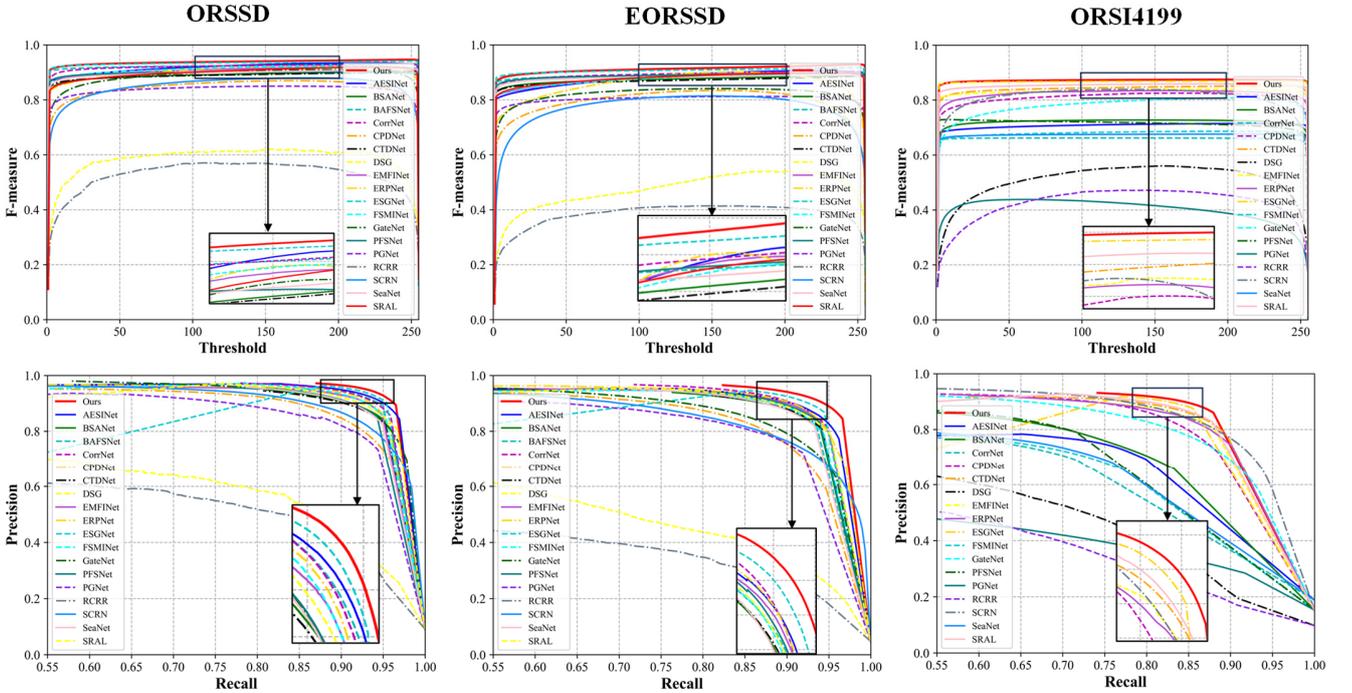

**Fig. 8.** Quantitative comparisons of different saliency models. The first and second rows are F-measure and P-R curves, respectively. *Please zoom-in for the best view.*

Overall, our method demonstrates strong competitiveness and superiority across the three test datasets. Specifically, as shown in Table I, on the ORSSD dataset, our method slightly underperforms ESGNet in $\mathcal{M}$, $E_\xi^{max}$ and $E_\xi^{mean}$. However, it outperforms ESGNet on other metrics ($S_\alpha$:0.9532 (Ours) vs. 0.9480 (ESGNet); $F_\beta^{max}$:0.9274 (Ours) vs. 0.9220 (ESGNet); $F_\beta^{mean}$: 0.9154 (Ours) vs. 0.9113 (ESGNet); $F_\beta^{adp}$: 0.9078 (Ours) vs. 0.9076 (ESGNet); $E_\xi^{adp}$: 0.9820 (Ours) vs. 0.9661 (ESGNet)). As shown in Table II, on the EORSSD dataset, our method's performance is slightly below BANet in the $F_\beta^{max}$, slightly below IP2GRNet in $F_\beta^{mean}$, and slightly below ESGNet in $F_\beta^{adp}$. However, overall, our method still maintains superiority. As shown in Table III, on the challenging ORSI4199 dataset, our method outperforms GLGCNet in nearly all metrics except for a marginal 0.04% lower score in $E_\xi^{mean}$ ($\mathcal{M}$:0.0268 (Ours) vs. 0.0274 (GLGCNet); $S_\alpha$: 0.8884 (Ours) vs. 0.8839 (GLGCNet); $F_\beta^{max}$: 0.8845 (Ours) vs. 0.8808 (GLGCNet); $F_\beta^{mean}$: 0.8787 (Ours) vs. 0.8712 (GLGCNet);

$F_\beta^{adp}$: 0.8764 (Ours) vs. 0.8672 (GLGCNet); $E_\xi^{max}$: 0.9518 (Ours) vs. 0.9508 (GLGCNet); $E_\xi^{adp}$: 0.9491 (Ours) vs. 0.9245 (GLGCNet)). Our method exhibits significant improvements compared to all other evaluated methods.

In summary, our method has demonstrated its superiority and effectiveness over these advanced methods through comprehensive qualitative analysis.

TABLE I
QUANTITATIVE COMPARISON RESULTS ON THE ORSSD DATASET. HERE, " ↑ " ( ↓ ) MEANS THAT THE LARGER (SMALLER) THE BETTER. THE BEST THREE RESULTS IN EACH ROW ARE MARKED IN **RED**, **BLUE**, AND **GREEN**, RESPECTIVELY

| Methods | Publication | ORSSD (200 images) | | | | | | | |
|---|---|---|---|---|---|---|---|---|---|
| | | $\mathcal{M} \downarrow$ | $S_\alpha \uparrow$ | $F_\beta^{max} \uparrow$ | $F_\beta^{mean} \uparrow$ | $F_\beta^{adp} \uparrow$ | $E_\xi^{max} \uparrow$ | $E_\xi^{mean} \uparrow$ | $E_\xi^{adp} \uparrow$ |
| Traditional methods | | | | | | | | | |
| DSG [23] | TIP 2017 | 0.1024 | 0.7186 | 0.6090 | 0.5682 | 0.5655 | 0.7849 | 0.7331 | 0.7538 |
| RCRR [21] | TIP 2018 | 0.1277 | 0.6849 | 0.5591 | 0.5127 | 0.4876 | 0.7651 | 0.7021 | 0.6957 |
| DL-based NSI-SOD methods | | | | | | | | | |
| SCRN [62] | ICCV 2019 | 0.0166 | 0.8951 | 0.8623 | 0.8296 | 0.7534 | 0.9523 | 0.9261 | 0.9079 |
| CPDNet [63] | CVPR 2019 | 0.0187 | 0.8954 | 0.8524 | 0.8243 | 0.7717 | 0.9438 | 0.9207 | 0.9211 |
| GateNet [64] | ECCV 2020 | 0.0131 | 0.9074 | 0.8913 | 0.8669 | 0.8065 | 0.9719 | 0.9562 | 0.9407 |
| CTDNet [65] | ACMM 2021 | 0.0138 | 0.9197 | 0.8853 | 0.8682 | 0.8561 | 0.9669 | 0.9620 | 0.9401 |
| PFSNet [66] | AAAI 2021 | 0.0142 | 0.9187 | 0.8832 | 0.8764 | 0.8716 | 0.9604 | 0.9528 | 0.9398 |
| BSANet [67] | AAAI 2022 | 0.0119 | 0.9170 | 0.8873 | 0.8690 | 0.8493 | 0.9635 | 0.9583 | 0.9555 |
| PGNet [68] | CVPR 2022 | 0.0225 | 0.8894 | 0.8338 | 0.8192 | 0.7921 | 0.9420 | 0.9218 | 0.9105 |
| DL-based ORSI-SOD methods | | | | | | | | | |
| FSMINet [31] | GRSL 2022 | 0.0101 | 0.9361 | 0.9042 | 0.8878 | 0.8710 | 0.9759 | 0.9671 | 0.9728 |
| ERPNet [29] | TCYB 2022 | 0.0114 | 0.9352 | 0.9036 | 0.8799 | 0.8392 | 0.9738 | 0.9604 | 0.9531 |
| ACCoNet [45] | TCYB 2022 | 0.0088 | 0.9437 | 0.9149 | 0.8971 | 0.8806 | 0.9796 | 0.9754 | 0.9721 |
| CorrNet [48] | TGRS 2022 | 0.0098 | 0.9380 | 0.9129 | 0.9002 | 0.8875 | 0.9790 | 0.9746 | 0.9756 |
| EMFINet [30] | TGRS 2022 | 0.0109 | 0.9373 | 0.9002 | 0.8856 | 0.8617 | 0.9737 | 0.9671 | 0.9663 |
| MJRBM [57] | TGRS 2022 | 0.0163 | 0.9204 | 0.8842 | 0.8566 | 0.8022 | 0.9623 | 0.9415 | 0.9328 |
| MCCNet [47] | TGRS 2022 | 0.0087 | 0.9437 | 0.9155 | 0.9054 | 0.8957 | 0.9800 | 0.9758 | 0.9735 |
| SRAL [70] | TGRS 2023 | 0.0105 | 0.9305 | 0.9167 | 0.8947 | 0.8612 | 0.9777 | 0.9674 | 0.9413 |
| SeaNet [41] | TGRS 2023 | 0.0105 | 0.9260 | 0.8942 | 0.8772 | 0.8625 | 0.9767 | 0.9721 | 0.9706 |
| BANet [71] | TGRS 2023 | 0.0097 | 0.9432 | 0.9205 | - | - | - | - | - |
| AESINet [40] | TGRS 2023 | 0.0085 | 0.9455 | 0.9160 | 0.8962 | 0.8640 | 0.9814 | 0.9741 | 0.9538 |
| CRNet [72] | TGRS 2023 | 0.0091 | 0.9389 | 0.9107 | 0.8951 | 0.8695 | 0.9793 | 0.9687 | 0.9711 |
| PCDNet [51] | TGRS 2023 | 0.0094 | 0.9444 | 0.9138 | 0.9014 | 0.8839 | 0.9809 | 0.9742 | 0.9731 |
| BAFSNet [37] | TGRS 2023 | 0.0079 | 0.9385 | 0.9136 | 0.9017 | 0.8982 | 0.9819 | 0.9785 | 0.9615 |
| ASTT [55] | TGRS 2023 | 0.0090 | 0.9350 | - | 0.9060 | - | - | 0.9700 | - |
| IP2GRNet [42] | JSTAR 2023 | 0.0090 | 0.9406 | - | **0.9121** | - | - | 0.9778 | - |
| CIFNet [50] | GRAL 2023 | 0.0084 | 0.9445 | 0.9183 | 0.9090 | **0.9028** | 0.9813 | 0.9773 | **0.9789** |
| ESGNet [32] | TCSVT 2023 | **0.0065** | 0.9484 | 0.9220 | 0.9113 | 0.9076 | **0.9858** | 0.9814 | 0.9661 |
| GLGCNet [34] | ISPRS 2023 | **0.0071** | 0.9488 | 0.9236 | 0.9054 | 0.8931 | **0.9864** | 0.9820 | 0.9800 |
| Ours | Submission | **0.0068** | 0.9532 | 0.9274 | 0.9154 | 0.9078 | 0.9847 | 0.9810 | 0.9820 |

TABLE II

QUANTITATIVE COMPARISON RESULTS ON THE EORSSD DATASET. HERE, " ↑ " ( ↓ ) MEANS THAT THE LARGER (SMALLER) THE BETTER. THE BEST THREE RESULTS IN EACH ROW ARE MARKED IN **RED**, **BLUE**, AND **GREEN**, RESPECTIVELY

| Methods | Publication | EORSSD (600 images) | | | | | | | |
|---|---|---|---|---|---|---|---|---|---|
| | | $\mathcal{M} \downarrow$ | $S_\alpha \uparrow$ | $F_\beta^{max} \uparrow$ | $F_\beta^{mean} \uparrow$ | $F_\beta^{adp} \uparrow$ | $E_\xi^{max} \uparrow$ | $E_\xi^{mean} \uparrow$ | $E_\xi^{adp} \uparrow$ |
| Traditional methods | | | | | | | | | |
| DSG | TIP 2017 | 0.1250 | 0.6426 | 0.5395 | 0.4734 | 0.4068 | 0.7286 | 0.6606 | 0.6191 |
| RCRR | TIP 2018 | 0.1647 | 0.6012 | 0.4142 | 0.3828 | 0.3476 | 0.6878 | 0.5967 | 0.5649 |
| DL-based NSI-SOD methods | | | | | | | | | |
| SCRN | ICCV 2019 | 0.0174 | 0.8499 | 0.8142 | 0.7682 | 0.5873 | 0.9366 | 0.8855 | 0.7740 |
| CPDNet | CVPR 2019 | 0.0111 | 0.8877 | 0.8344 | 0.7905 | 0.6843 | 0.9390 | 0.8980 | 0.8651 |
| GateNet | ECCV 2020 | 0.0130 | 0.8724 | 0.8409 | 0.8187 | 0.7153 | 0.9352 | 0.9146 | 0.8677 |
| CTDNet | ACMM 2021 | 0.0088 | 0.9195 | 0.8876 | 0.8664 | 0.8409 | 0.9646 | 0.9601 | 0.9216 |
| PFSNet | AAAI 2021 | 0.0079 | 0.9247 | 0.8975 | 0.8851 | 0.8656 | 0.9659 | 0.9610 | 0.9330 |
| BSANet | AAAI 2022 | 0.0080 | 0.9183 | 0.8876 | 0.8708 | 0.8399 | 0.9648 | 0.9595 | 0.9489 |
| PGNet | CVPR 2022 | 0.0153 | 0.8754 | 0.8124 | 0.8005 | 0.7631 | 0.9302 | 0.9171 | 0.8858 |
| DL-based ORSI-SOD methods | | | | | | | | | |
| FSMINet | GRSL 2022 | 0.0079 | 0.9256 | 0.8946 | 0.8699 | 0.8247 | 0.9667 | 0.9564 | 0.9490 |
| ERPNet | TCYB 2022 | 0.0082 | 0.9255 | 0.9013 | 0.8531 | 0.7392 | 0.9664 | 0.9366 | 0.8994 |
| ACCoNet | TCYB 2022 | 0.0074 | 0.9290 | 0.8837 | 0.8552 | 0.7969 | 0.9727 | 0.9653 | 0.9450 |
| CorrNet | TGRS 2022 | 0.0083 | 0.9298 | 0.9049 | 0.8890 | **0.8568** | 0.9696 | 0.9646 | 0.9593 |
| EMFINet | TGRS 2022 | 0.0084 | 0.9299 | 0.8990 | 0.8751 | 0.8231 | 0.9706 | 0.9601 | 0.9501 |
| MJRBM | TGRS 2022 | 0.0099 | 0.9197 | 0.8656 | 0.8239 | 0.7066 | 0.9646 | 0.9350 | 0.8897 |
| MCCNet | TGRS 2022 | 0.0066 | 0.9327 | 0.8904 | 0.8604 | 0.8137 | 0.9755 | 0.9685 | 0.9538 |
| SRAL | TGRS 2023 | 0.0067 | 0.9234 | 0.8964 | 0.8743 | 0.8255 | 0.9703 | 0.9608 | 0.9203 |
| SeaNet | TGRS 2023 | 0.0073 | 0.9208 | 0.8649 | 0.8519 | 0.8304 | 0.9710 | 0.9650 | **0.9614** |
| BANet | TGRS 2023 | 0.0089 | 0.9292 | **0.9055** | - | - | - | - | - |
| AESINet | TGRS 2023 | 0.0064 | 0.9347 | 0.8792 | 0.8496 | 0.7909 | 0.9757 | 0.9646 | 0.9297 |
| CRNet | TGRS 2023 | 0.0063 | 0.9370 | 0.8873 | 0.8653 | 0.8140 | 0.9743 | 0.9618 | 0.9563 |
| PCDNet | TGRS 2023 | 0.0069 | 0.9346 | 0.8830 | 0.8590 | 0.8088 | 0.9744 | 0.9645 | 0.9507 |
| BAFSNet | TGRS 2023 | 0.0060 | 0.9287 | 0.8708 | 0.8583 | 0.8477 | 0.9772 | 0.9734 | 0.9441 |
| ASTT | TGRS 2023 | 0.0060 | 0.9250 | - | 0.8740 | - | - | 0.9580 | - |
| IP2GRNet | JSTAR 2023 | 0.0071 | 0.9248 | - | **0.8962** | - | - | **0.9756** | - |
| CIFNet | GRAL 2023 | 0.0061 | 0.9342 | 0.8843 | 0.8624 | 0.8249 | 0.9763 | 0.9709 | 0.9569 |
| ESGNet | TCSVT 2023 | **0.0051** | 0.9373 | 0.8907 | 0.8752 | **0.8632** | **0.9781** | 0.9731 | 0.9461 |
| GLGCNet | ISPRS 2023 | **0.0055** | **0.9375** | 0.8924 | 0.8714 | 0.8499 | **0.9803** | **0.9751** | **0.9701** |
| Ours | Submission | **0.0046** | **0.9433** | **0.9008** | **0.8811** | **0.8548** | **0.9823** | **0.9773** | **0.9732** |

*2) Qualitative analysis:* To enhance the visualization of our method's efficacy in ORSI-SOD, we selected five representative and challenging scenarios for qualitative analysis, as depicted in Fig. 9. The five scenarios are summarized as follows: 1) three cases of airplanes: general airplane, airplane with shadows, and multiple airplanes; 2) three cases of buildings: general building, multiple buildings, and incomplete buildings; 3) three cases of cars: multiple cars, incomplete cars, and cars with shadows; 4) two cases of lakes: lake with low contrast and narrow lake; 5) two cases of rivers: river with low contrast and river with irregular topology. For each scenario, we display the original images alongside the ground truth and the saliency maps produced by 17 different methods. The comparison clearly demonstrates our method's superior accuracy and comprehensiveness relative to the others, showcasing its effectiveness in addressing various ORSI-SOD challenges.

TABLE III
QUANTITATIVE COMPARISON RESULTS ON THE ORSI-4199 DATASET. HERE, "↑" (↓) MEANS THAT THE LARGER (SMALLER) THE BETTER. THE BEST THREE RESULTS IN EACH ROW ARE MARKED IN **RED**, **BLUE**, AND **GREEN**, RESPECTIVELY

| Methods | Publication | ORSI-4199 (2199 images) | | | | | | | |
| --- | --- | --- | --- | --- | --- | --- | --- | --- | --- |
| | | $\mathcal{M} \downarrow$ | $S_\alpha \uparrow$ | $F_\beta^{max} \uparrow$ | $F_\beta^{mean} \uparrow$ | $F_\beta^{adp} \uparrow$ | $E_\xi^{max} \uparrow$ | $E_\xi^{mean} \uparrow$ | $E_\xi^{adp} \uparrow$ |
| Traditional methods | | | | | | | | | |
| DSG | TIP 2017 | 0.1294 | 0.6971 | 0.6270 | 0.5560 | 0.5477 | 0.7833 | 0.7153 | 0.7320 |
| RCRR | TIP 2018 | 0.1637 | 0.6491 | 0.5480 | 0.4924 | 0.4806 | 0.7192 | 0.6579 | 0.6730 |
| DL-based NSI-SOD methods | | | | | | | | | |
| SCRN | ICCV 2019 | 0.0294 | 0.8689 | 0.8484 | 0.8153 | 0.7795 | 0.9361 | 0.8953 | 0.8940 |
| CPDNet | CVPR 2019 | **0.0281** | 0.8634 | 0.8343 | 0.8155 | 0.7929 | 0.9244 | 0.9071 | 0.9139 |
| GateNet | ECCV 2020 | 0.0378 | 0.8425 | 0.8169 | 0.7852 | 0.7370 | 0.9133 | 0.8925 | 0.8635 |
| CTDNet | ACMM 2021 | 0.0312 | 0.8727 | 0.8617 | 0.8487 | 0.8177 | 0.9416 | 0.9352 | 0.9091 |
| PFSNet | AAAI 2021 | 0.0791 | 0.7596 | 0.7372 | 0.7227 | 0.7397 | 0.8557 | 0.8170 | 0.8406 |
| BSANet | AAAI 2022 | 0.0471 | 0.7965 | 0.7342 | 0.7259 | 0.7312 | 0.8633 | 0.8423 | 0.8744 |
| PGNet | CVPR 2022 | 0.2177 | 0.5683 | 0.4371 | 0.4009 | 0.3990 | 0.6094 | 0.5723 | 0.6401 |
| DL-based ORSI-SOD methods | | | | | | | | | |
| FSMINet | GRSL 2022 | 0.0788 | 0.7558 | 0.6946 | 0.6856 | 0.6775 | 0.8383 | 0.8315 | 0.8370 |
| ERPNet | TCYB 2022 | 0.0321 | 0.8635 | 0.8342 | 0.8223 | 0.8060 | 0.9296 | 0.9164 | 0.8849 |
| ACCoNet | TCYB 2022 | 0.0314 | 0.8775 | 0.8686 | 0.8620 | 0.8581 | 0.9412 | 0.9342 | 0.9167 |
| CorrNet | TGRS 2022 | 0.0366 | 0.8623 | 0.8560 | 0.8513 | 0.8534 | 0.9330 | 0.9206 | 0.9142 |
| EMFINet | TGRS 2022 | 0.0376 | 0.8664 | 0.8486 | 0.8389 | 0.8319 | 0.9260 | 0.9170 | 0.9236 |
| MJRBM | TGRS 2022 | 0.0374 | 0.8593 | 0.8493 | 0.8309 | 0.7995 | 0.9311 | 0.9102 | 0.8891 |
| MCCNet | TGRS 2022 | 0.0316 | 0.8746 | 0.8690 | 0.8630 | 0.8592 | 0.9413 | 0.9348 | 0.9182 |
| SRAL | TGRS 2023 | 0.0321 | 0.8735 | 0.8576 | 0.8482 | 0.8119 | 0.9442 | 0.9341 | 0.9181 |
| SeaNet | TGRS 2023 | 0.0308 | 0.8722 | 0.8653 | 0.8591 | 0.8556 | 0.9426 | 0.9362 | **0.9394** |
| BANet | TGRS 2023 | 0.0314 | 0.8767 | 0.8576 | - | - | - | - | - |
| AESINet | TGRS 2023 | 0.0305 | 0.8755 | 0.8726 | 0.8626 | 0.8246 | 0.9459 | 0.9372 | 0.9196 |
| CRNet | TGRS 2023 | 0.0335 | 0.8669 | 0.8669 | - | - | 0.9176 | - | - |
| ESGNet | TCSVT 2023 | 0.0290 | **0.8795** | **0.8788** | **0.8733** | **0.8740** | 0.9495 | **0.9438** | **0.9484** |
| GLGCNet | ISPRS 2023 | **0.0274** | **0.8839** | **0.8808** | **0.8712** | **0.8672** | **0.9508** | **0.9469** | 0.9245 |
| Ours | Submission | **0.0268** | **0.8884** | **0.8845** | **0.8787** | **0.8764** | **0.9518** | **0.9465** | **0.9491** |

Specifically, in the evaluation of challenging scenarios, the two traditional SOD methods underperformed significantly, apart from rivers with irregular topologies. The four NSI-SOD methods displayed potential in some instance but generally struggled across a broad spectrum of challenging ORSI scenarios. Notably, as depicted in the first two rows of Fig. 9, columns 5 to 6, these methods accurately identify a general airplane but their detection accuracy for airplanes with shadows are impacted by shadows. The 10 specialized ORSI-SOD methods are capable of locating salient objects. However, their performance wane in precisely delineating boundary details and achieving completeness. For example, in row 10 and row 12, which presented scenario with low contrast, the majority of ORSI-SOD methods struggle with background noise are unable to fully outline the lakes' shapes. Even when the outer contours of rivers are recognized, the subtle contrast between the rivers and their backgrounds often result in missed internal features. Conversely, our method stands out for its precise detection of salient objects and discernment of finer boundaries across these challenging scenarios. It showcases notable superiority, especially in scenes featuring incomplete cars, incomplete buildings and narrow lake, as shown in rows 6, 8 and 11, clearly outperforming the others. Overall, our method not only accurately identifies salient objects but also captures intricate details in complex scenarios, demonstrating its exceptional adaptability and robustness.

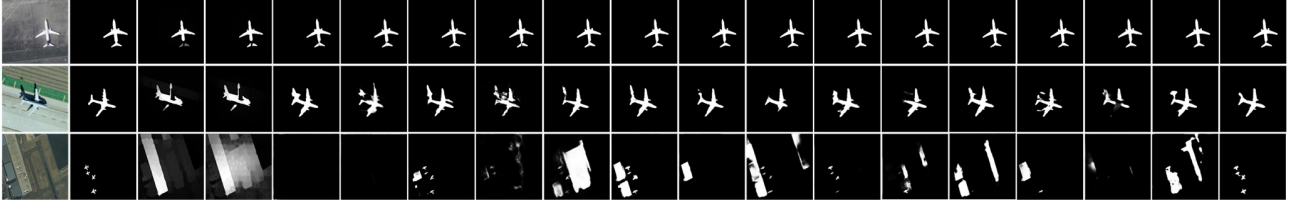
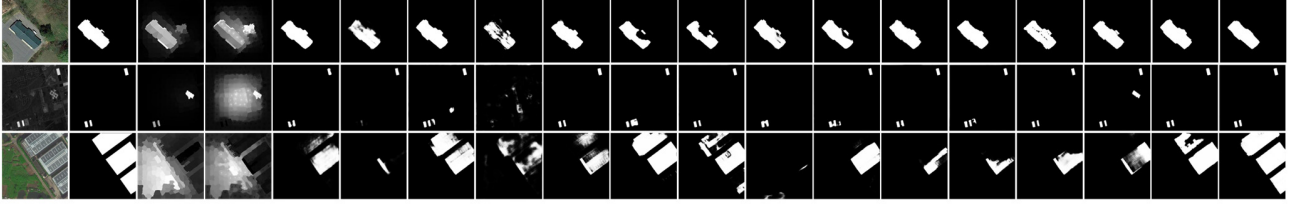
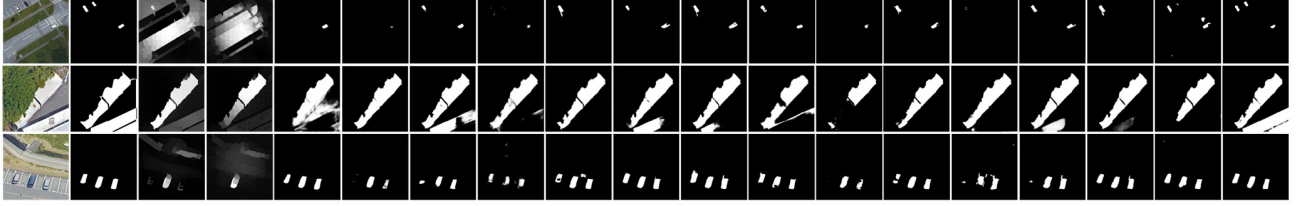
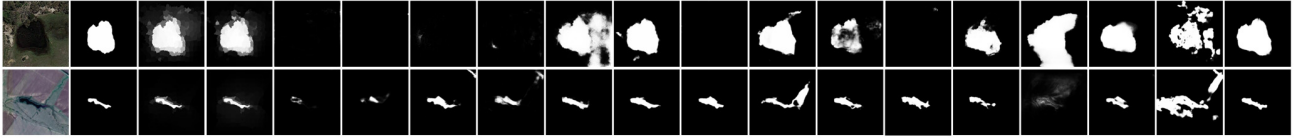
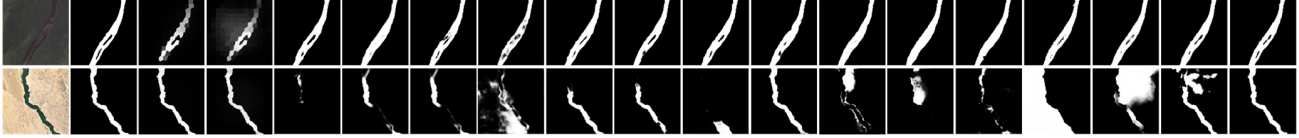

**Fig. 9.** Qualitative analysis of different saliency methods. (a) Optical RSIs. (b) GT. (c) DSG. (d) RCRR. (e) BSANet. (f) CPDNet. (g) CTDNet. (h) GateNet. (i) ACCoNet. (j) AESINet. (k) MCCNet. (l) FSMINet. (m) SRAL. (n) CorrNet. (o) SeaNet. (p) EMFINet. (q) ERPNet. (r) ESGNet. (s) Ours. *Please zoom-in for the best view.*

TABLE IV
TRANSFERABILITY VERIFICATION ON THE RSISOD DATASET. HERE, "↑" (↓) MEANS THAT THE LARGER (SMALLER) THE BETTER. THE BEST THREE RESULTS IN EACH ROW ARE MARKED IN **RED**, **BLUE**, AND **GREEN**, RESPECTIVELY.

| Metrics | $\mathcal{M}\downarrow$ | $S_\alpha\uparrow$ | $F_\beta^{max}\uparrow$ | $F_\beta^{mean}\uparrow$ | $F_\beta^{adp}\uparrow$ | $E_\xi^{max}\uparrow$ | $E_\xi^{mean}\uparrow$ | $E_\xi^{adp}\uparrow$ |
|---|---|---|---|---|---|---|---|---|
| ACCoNet | **0.0988** | **0.7582** | 0.6977 | 0.6848 | 0.6642 | **0.8065** | **0.8001** | 0.7671 |
| AESINet | **0.0926** | **0.7652** | **0.7145** | **0.7047** | 0.6438 | **0.8179** | **0.8123** | **0.7794** |
| CorrNet | 0.1054 | 0.7322 | 0.6620 | 0.6590 | 0.6629 | 0.7890 | 0.7785 | 0.7696 |
| FMSINet | 0.1047 | 0.7255 | 0.6415 | 0.6338 | 0.6073 | 0.7801 | 0.7703 | 0.7634 |
| MCCNet | 0.1117 | 0.7341 | 0.6687 | 0.6636 | 0.6606 | 0.7863 | 0.7830 | 0.7604 |
| MJRBM | 0.1088 | 0.7306 | 0.6619 | 0.6450 | 0.6348 | 0.7901 | 0.7681 | 0.7584 |
| SeaNet | 0.1014 | 0.7410 | 0.6761 | 0.6719 | **0.6723** | 0.7958 | 0.7911 | **0.7783** |
| Ours | **0.0912** | **0.7741** | **0.7299** | **0.7223** | **0.7108** | **0.8254** | **0.8210** | 0.7726 |

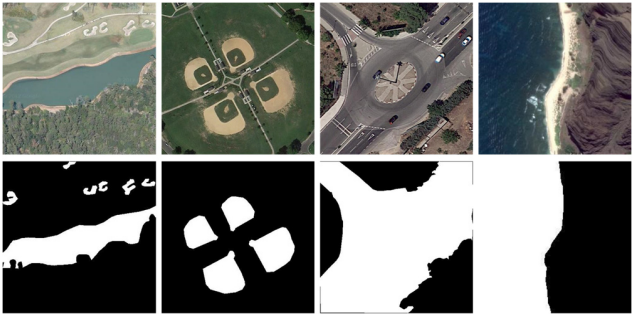

**Fig. 10.** Some representative samples: Optical remote sensing images (up) and GT (down).

*C. The Validation of Model Transferability*

To assess the model's transferability, we engaged the larger ORSI-SOD dataset in a transferability study. The RSISOD dataset, recently compiled by Zheng et al. [71], stands out for its comprehensive inclusion of real-world scenarios and an expanded spectrum of salient object characteristics. Fig. 10 showcases select images from the RSISOD dataset, alongside their pixel-level GT. It consists of 3783 images for training and 1270 images with their GTs for testing. For a more rigorous transferability assessment, we merged the training and test sets into a unified test set, culminating in 5053 images for the transferability experiment. To maintain experiment fairness, the model was tested using weights trained on the ORSI-4199 dataset. Additionally, for this experiment, we selected ORSI-SOD methods for which authors provided source codes and training weights tailored to the ORSI-4199 dataset, ensuring a consistent basis for comparison.

Table IV delineates the performance outcomes of various ORSI-SOD methods on the RSISOD dataset. In the table, the $E_\xi^{adp}$ of our method is inferior to AESINet and SeaNet by 0.68% and 0.57% respectively. However, our method surpasses them in other evaluation metrics ($\mathcal{M}$: 0.0912 (Ours) vs. 0.0926 (AESINet) vs. 0.1014 (SeaNet); $S_\alpha$: 0.7741 (Ours) vs. 0.7652 (AESINet) vs. 0.7410 (SeaNet); $F_\beta^{max}$: 0.7299 (Ours) vs. 0.7145 (AESINet) vs. 0.6761 (SeaNet); $F_\beta^{mean}$: 0.7223 (Ours) vs. 0.7047 (AESINet) vs. 0.6761 (SeaNet); $F_\beta^{adp}$: 0.7108 (Ours) vs. 0.6438 (AESINet) vs. 0.6723 (SeaNet); $E_\xi^{max}$: 0.8254 (Ours) vs. 0.8179 (AESINet) vs. 0.7958 (SeaNet); $E_\xi^{mean}$: 0.8210 (Ours) vs. 0.8123 (AESINet) vs. 0.7911 (SeaNet)). These results underscore our method's superior generalization ability over other ORSI-SOD approaches, reinforcing its superiority in the domain.

*D. Computational Efficiency Analysis*

To evaluate the computational efficiency of our method, we benchmarked it against others based on network parameters (Param), floating-point operations (FLOPs), and frames per second (FPS), with the findings illustrated in Fig. 11. In Fig. 11, red bars represent Param, blue bars represent FLOPs, and the black line represents FPS. The results clearly indicate that ACCoNet has the highest Param, yet its FPS performance is not optimal. MJBRM exhibits high FLOPs, indicating substantial computational demand during execution, but it has a lower FPS. Notably, FSMINet and SeaNet, which are tailored for lightweight applications, register the minimal network parameters and FLOPs, as well as the highest FPS. Compared to most other methods, our model maintains a better balance in network parameters and FLOPs while excelling in various performance metrics. Additionally, our accuracy on three datasets significantly surpasses theirs. Combined with the results presented in Table I, Table II and Table III, our method maintains favorable network parameters, FLOPs, and FPS, while ensuring high-precision outcomes. In summary, our method surpasses the other methods in these aspects. In conclusion, our method exceeds the performance of competing methods across these key metrics.

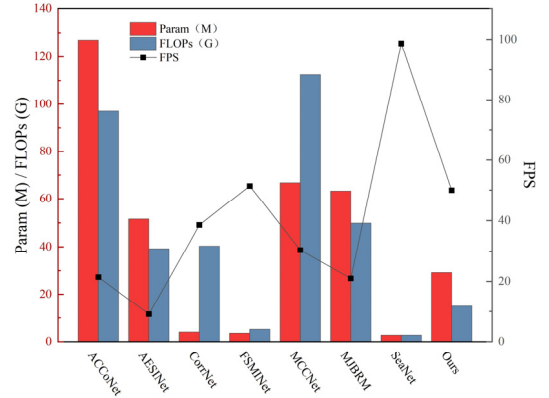

**Fig. 11.** Computational efficiency analysis of different salience methods.

*E. Ablation Analysis*

In this section, we validate the effectiveness of key components within the network through extensive ablation studies conducted on the ORSI4199 dataset.

*1) Ablation Study with Different Components:* To verify the effectiveness of the EFABA and GDAL modules, we designed three variants by sequentially adding modules to the baseline: 1) Baseline, where we use PVT-V2-B2 as the baseline; 2) Baseline + EFABA; 3) Baseline + GDAL. The quantitative analysis results are shown in Table V. In table, adding our proposed EFABA module and GDAL module to the baseline separately results in a significant improvement in accuracy. This confirms the effectiveness of the EFABA and GDAL modules within the network. However, the accuracy of these variants is significantly lower than that of the complete LBA-MCNet, which demonstrates that both modules are essential for effective salient object detection. This further proves the superiority of our LBA-MCNet. Additionally, we compare the saliency maps of different variants in Fig. 12. From the figure, it is evident that the Baseline is susceptible to errors and omissions due to interference from complex backgrounds. The incorporation of the EFABA module, which captures edge details and perceives local context, enables effective detection of salient objects while being more sensitive to boundary information. The GDAL module is designed to model the global context information at the image level, effectively suppressing interference from complex backgrounds and accurately identifying salient objects. The figure shows that adding each module can effectively detect salient objects, but there are still certain limitations. Our method combines the two, judiciously utilizing the strengths of both modules to correctly and completely detect salient objects while overcoming complex background interference.

TABLE V
ABLATION EXPERIMENTAL RESULTS OF DIFFERENT COMPONENTS. HERE, "↑"(↓) MEANS THAT THE LARGER (SMALLER) THE BETTER. THE BEST RESULT IS **BOLD**

| No. | Different components | | | ORSI4199 | | |
|---|---|---|---|---|---|---|
| | Baseline | EFABA | GDAL | $\mathcal{M} \downarrow$ | $S_\alpha \uparrow$ | $F_\beta^{max} \uparrow$ |
| 1 | ✓ | | | .0367 | .8595 | .8467 |
| 2 | ✓ | ✓ | | .0347 | .8723 | .8619 |
| 3 | ✓ | | ✓ | .0319 | .8749 | .8654 |
| 4 | ✓ | ✓ | ✓ | **.0268** | **.8884** | **.8845** |

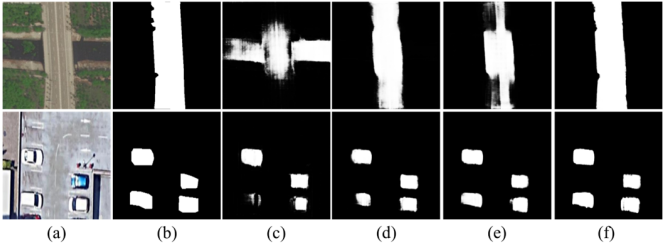

**Fig. 12.** Visualization of ablation experiments between different modules. (a)Optical RISs. (b) GT. (c) Baseline. (d) Baseline + EFABA. (e) Baseline + GDAL. (f) Baseline + EFABA + GDAL (Ours).

*2) Ablation Study with EFABA module:* The EFABA module consists of two sub-modules: the ECD module and the FABA module. We designed four variants for evaluation: 1) Baseline; 2) w/ FABA: incorporating the FABA module; 3) w/ EFABA: incorporating the EFABA module, the same as No.2 in Table III; 4) w/o ECD: removing only the ECD module. The quantitative comparison results are shown in Table VI. From the table, it is evident that adding the FABA sub-module to the Baseline significantly improves accuracy, indicating that the FABA module effectively balances detail features with local context at different levels. The accuracy of adding only the FABA module is less than that of adding the EFABA module, and removing the ECD module from the network results in a decrease in overall accuracy, which demonstrates that the ECD module can effectively capture edge features in the data. Overall, both the ECD and FABA modules are effective in detecting salient objects.

To further understand the contributions of the EFABA module, we conduct visual analyses on the ECD module and the FABA module separately. ECD module is made to discern basic edge features. Illustrated in Fig. 13, ($a_1$) and ($a_2$) showcase the feature outputs from the encoder's initial two layers, while E symbolizes the output edges, denoted as $e^{att}$ in section III(B)1). The visual analysis indicates that the first two layers of the encoder are abundant in boundary information, and ECD module could capture the edges of salient objects accurately. In particular, compared with the model without ECD (w/o ECD), the model with ECD module (w/ECD) exhibits a marked improvement in capturing the boundaries of rivers both distinctly and accurately.

TABLE VI
ABLATION EXPERIMENTAL RESULTS OF EFABA MODULE. HERE, "↑"(↓) MEANS THAT THE LARGER (SMALLER) THE BETTER. THE BEST RESULT IS **BOLD**

| No. | Different components | ORSI4199 | | |
|---|---|---|---|---|
| | | $\mathcal{M} \downarrow$ | $S_\alpha \uparrow$ | $F_\beta^{mean} \uparrow$ |
| 1 | Baseline | .0367 | .8595 | .8467 |
| 2 | w/ FABA | .0354 | .8631 | .8544 |
| 3 | w/ EFABA | .0347 | .8723 | .8619 |
| 4 | w/o ECD | .0292 | .8831 | .8711 |
| 5 | Ours | **.0268** | **.8884** | **.8845** |

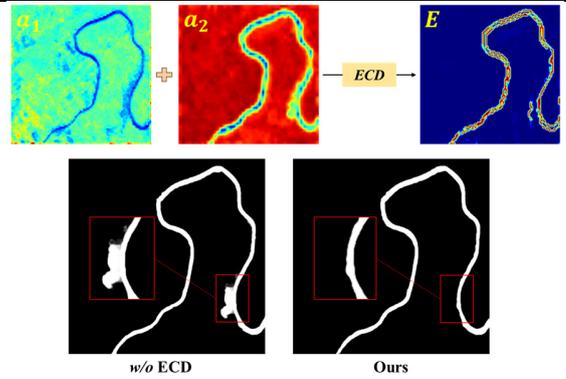

**Fig. 13.** Visualization of feature maps and results from ECD module. $a_1$ and $a_2$ represent encoder's first two layers, E represents the output edge.

Furthermore, the FABA module is specifically engineered to balance detailed features with the local context throughout the encoder's various layers. As illustrated in Fig. 14, ($a_1$) - ($a_3$) display the encoder's first to third layers, ($b_1$) - ($b_3$) showcase the results processed by FABA module. It is obvious that the clarity of boundary position information diminishes through the encoder's forward propagation, particularly in ($a_3$), where much of the edge information is lost. After being processed by the FABA module, salient objects emerge with greater clarity and spatial details are more effectively conserved throughout the propagation process, thanks to the effective use of features across all layers. Compared with ($a_1$), in ($b_1$), the definition and precision of edge information are markedly enhanced. ($a_2$) is predominantly oriented towards capturing background details, while ($b_2$) shift focus towards salient objects, emphasizing the intricate details of edge locations with greater clarity. The impact of the FABA module is particularly pronounced in ($b_3$), where salient objects are more prominently highlighted and their spatial positions are depicted with greater accuracy. Overall, the visualized feature maps serve as a compelling illustration of the EFABA module's significant contributions.

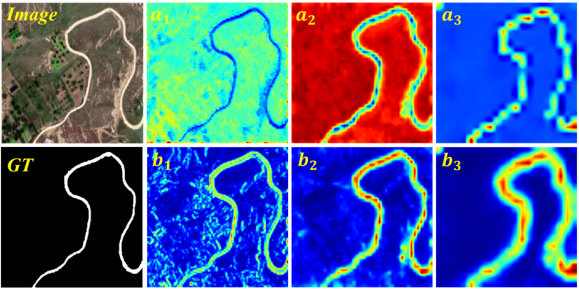

**Fig. 14.** Visualization of feature maps before and after EFABA module. $(a_1)$ - $(a_3)$ represent the encoder's first to third layers, $(b_1)$ - $(b_3)$ represent the results processed by FABA module.

Finally, we select the latest three boundary-oriented methods for edge detection comparison, including SeaNet, AESINet, and ESGNet, as shown in Fig. 15. In Fig. 15, building boundaries are partially obscured by trees, as highlighted in red boxes. These three methods that focus solely on either spatial or channel attention face limitations in adequately addressing edge details and local semantics. This singular focus results in a lack of comprehensive attention to the intricate balance required between the sharpness of edge features and the richness of the local context, leading to the manifestation of unclear building boundaries in their outputs. In contrast, EFABA effectively implements the balance between the detailed features of edges and local semantics. This module minimizes background interference on the edges of buildings, leading to the production of clearer and more defined boundaries for salient objects. The enhanced clarity and precision of object boundaries achieved by the EFABA more accurately reflect the ground truth.

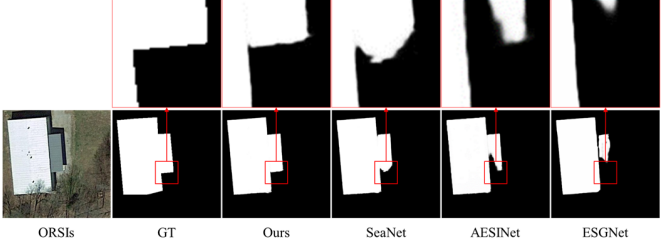

**Fig. 15.** Visual comparisons with 3 representative methods for edges.

*3) Ablation Study with GDAL Module:* The GDAL module is divided into the EAL module and IAL module. We designed four variants to validate the effectiveness of the GDAL module: 1) Baseline; 2) Baseline+EAL; 3) Baseline+IAL; 4) Baseline+EAL+IAL, which is the same as No.3 in Table V. From the quantitative analysis results in Table VII, it is evident that adding either the EAL or IAL module separately improves network accuracy, but neither achieves the high accuracy provided by incorporating the entire GDAL module. Both the EAL and IAL modules can effectively model global contextual information, but combining these modules leverages their respective strengths more effectively, enhancing the utilization of global context and promoting the integration of global and local information.

TABLE VII
ABLATION EXPERIMENTAL RESULTS OF GDAL MODULE. HERE, "↑"(↓) MEANS THAT THE LARGER (SMALLER) THE BETTER. THE BEST RESULT IS **BOLD**

| No. | Baseline | EAL | IAL | ORSI4199 | | |
|---|---|---|---|---|---|---|
| | | | | $\mathcal{M}\downarrow$ | $S_\alpha\uparrow$ | $F_\beta^{max}\uparrow$ |
| 1 | ✓ | | | .0367 | .8595 | .8467 |
| 2 | ✓ | ✓ | | .0351 | .8636 | .8523 |
| 3 | ✓ | | ✓ | .0345 | .8677 | .8580 |
| 4 | ✓ | ✓ | ✓ | **.0319** | **.8749** | **.8654** |

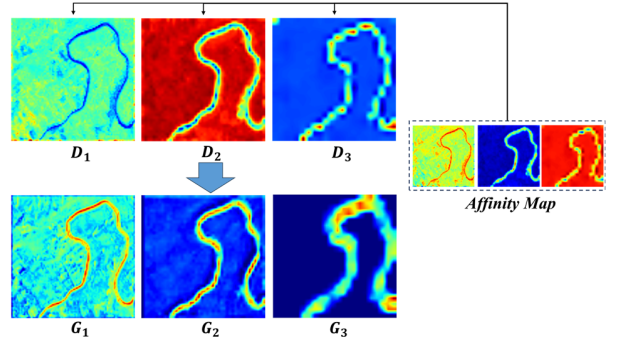

**Fig. 16.** Visualization of feature maps before and after GDAL module. $(D_1)$ - $(D_3)$ represent the encoder's first three layers, $(G_1)$ - $(G_3)$ represent the feature maps processed by GDAL module.

As detailed in section III(C), the primary function of GDAL module is to model the image-level global context. It leverages the encoder's final layer, which harbors the richest semantic information, to create affinity maps. These maps are subsequently reintegrated into the encoder's initial three layers, enhancing the network's ability to learn hierarchical features and effectively grasp the image-level global context. Illustrated in Fig. 16, $(D_1)$ - $(D_3)$ display the encoder's first three layers, and $(G_1)$ - $(G_3)$ represent the feature maps processed by GDAL module. Specifically, the global context association map (Affinity Map) generated by the encoder's last layer are imbued with rich semantic information. Observing from $(D_1)$ to $(D_3)$, it's clear that as the encoder advances, the feature maps become enriched with global information, albeit with a reduction in resolution and an expansion of the receptive field per pixel. The reintroduction of these affinity maps into the encoder's initial three layers enables the network to assimilate a more substantial volume of global information. For ($D_1$), the GDAL module significantly bolsters the network's capacity to recognize saliency features, as evidenced by the salient objects' emphasized areas in ($G_1$). This enhancement is credited to the GDAL module's capacity to prioritize global features, leveraging the SE attention mechanism to sharpen the model's focus on feature channels that are particularly discriminative for identifying various object types. In $(D_2)$, the encoder initially diverts its attention towards the image background rather than salient objects. However, as shown in ($G_2$), the semantic information pertaining to salient objects is significantly bolstered post-

GDAL module processing. In ($D_3$), while the semantic details of salient objects initially lack prominence, the GDAL's processing markedly amplifies the encoder's attention towards pertinent information in ($G_3$). Overall, the pivotal role of the GDAL module in enhancing model performance is vividly demonstrated through these visualized feature maps.

*F. Failure Cases and Limitations*

While the experiments conducted have comprehensively affirmed the efficacy of our approach, LBA-MCNet, there remain scenarios where it faces challenges. Fig. 17 showcases some illustrative failure cases, which essentially fall into two primary categories: 1) multiple salient objects, as depicted in the first two columns. It's significant to acknowledge that while the training set of the original dataset includes images of these objects, which might be prominent in various scenes, the dataset seldom features images with two prominent objects possessing distinct attributes. This scarcity can contribute to a discrepancy between the results produced by our method and the actual ground truth values. 2) similar object and background colors, as illustrated in the last two columns. The close resemblance in colors between background and salient objects presents a significant challenge in completely mitigating interference. Moreover, as Table III illustrates, our method demonstrates substantial benefits when applied to larger and more complex datasets, such as ORSI-4199. This success can be chiefly ascribed to our approach's proficiency in harmonizing boundary details with local semantics, alongside its effectiveness in modeling the image-level global context. However, the constraints posed by the limited size and diversity of datasets might restrict our ability to fully address a broad spectrum of scenarios and variations, as well as to adequately capture essential semantic information and spatial details.

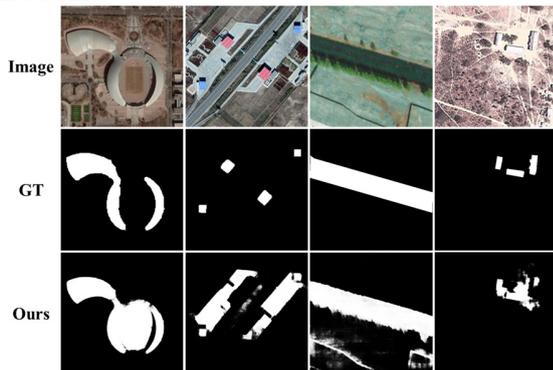

**Fig. 17.** Some representative fail cases.

## V. CONCLUSION

In this paper, we delve into the significance of boundary and contextual features for SOD and present LBA-MCNet, a cutting-edge approach tailored for ORSI-SOD. Leveraging a transformer as our foundational backbone for feature extraction, LBA-MCNet integrates two pivotal modules: EFABA and GDAL. The EFABA module is adept at identifying edge details within shallow features, significantly improving the model's capacity to locate salient objects. It synergizes these details with an attention mechanism that adeptly captures the local context, all the while maintaining essential semantics. This strategic combination ensures a harmonious equilibrium between spatial precision and local semantic richness across the network's various layers. Moreover, the GDAL module concentrates on leveraging deep features to construct image-level global context association graphs through explicit affinity learning. It further enriches the model by encoding global information via implicit affinity learning, efficiently recognizing global patterns within images. This module also fosters a more cohesive expression of multi-layer features by applying deep supervision within the decoding layer. An extensive comparative analysis, involving 28 different methods across three publicly accessible datasets, underscores LBA-MCNet's exceptional performance and its distinguished place within the domain of ORSI-SOD, confirming the method's effectiveness and superior capability. Additionally, we plan to improve EFABA and GDAL to make them plug-and-play modules, allowing for more effective application to other salient object detection tasks in the future.